\documentclass[letterpaper, 10 pt, conference]{ieeeconf}   

\IEEEoverridecommandlockouts                              
\usepackage{graphicx}
\usepackage{subcaption,booktabs}
\usepackage{float}
\usepackage{comment}
\usepackage{color}
\usepackage[normalem]{ulem}
\graphicspath{{Images/}}
\usepackage{subcaption}
\usepackage{graphicx}

\bstctlcite{IEEEexample:BSTcontrol}

\title{\LARGE \bf
A Task Allocation Approach for Human-Robot Collaboration in Product Defects Inspection Scenarios}

\author{Hossein Karami$^{1}$, Kourosh Darvish$^{2}$, Fulvio Mastrogiovanni$^{1}$
\thanks{${^1}$ Department of Informatics, Bioengineering,
Robotics, and Systems Engineering, University of Genoa, Via Opera Pia 13,
16145, Genoa, Italy.}%
\thanks{${^2}$ Dynamic Interaction Control, Istituto Italiano di Tecnologia, Genova, Italy.}%
\thanks{Corresponding author’s email: hossein.karami@edu.unige.it.}%
}

\begin{document}

\maketitle
\thispagestyle{empty}
\pagestyle{empty}

\begin{abstract}

The presence and coexistence of human operators and collaborative robots in shop-floor environments raises the need for assigning tasks to either operators or robots, or both. 
Depending on task characteristics, operator capabilities and the involved robot functionalities, it is of the utmost importance to design strategies allowing for the concurrent and/or sequential allocation of tasks related to object manipulation and  assembly. 
In this paper, we extend the \textsc{FlexHRC} framework presented in \cite{darvish2018flexible} to allow a human operator to interact with multiple, heterogeneous robots at the same time in order to jointly carry out a given task. 
The extended \textsc{FlexHRC} framework leverages a concurrent and sequential task representation framework to allocate tasks to either operators or robots as part of a dynamic collaboration process.
In particular, we focus on a use case related to the inspection of product defects, which involves a human operator, a dual-arm Baxter manipulator from Rethink Robotics and a Kuka youBot mobile manipulator.

\end{abstract}

\section{Introduction}

Robots are increasingly adopted in industrial environments to carry out dangerous, repetitive, or stressful tasks.
The introduction of robots in production lines has improved a number of key performance indicators, and has addressed a market-driven goods growing demand with quality products \cite{esmaeilian2016evolution}.
However, due to well-known limitations of robot perceptual, cognitive, and reasoning capabilities, certain tasks, which are difficult to model or require a higher-level of awareness because they cannot be easily modelled nor formalised, are still better handled by human operators.
The introduction of collaborative robots (nowadays referred to as \textit{cobots}) in recent years has contributed to relax those limitations, and implicitly promoted human working conditions \cite{kock2011robot}. 
Among the tasks typically considered stressful, \textit{quality control} and \textit{defects inspection} play a key role in defining the quality of a finished or semi-finished product.
Currently, trained and expert personnel is tasked with establishing benchmarks and examining products quality, which require prolonged focus and continuous attention.
In this work, we argue that the collaboration between an experienced human operator and a robot may lead to higher rates in defects spotting, overall productivity, and safety \cite{de2008atlas, lasota2017survey}.

Human-robot collaboration (HRC) is defined as the purposeful interaction among humans and robots in a shared space, and it is aimed at a common goal. 
A natural collaboration requires a robot to perceive and correctly interpret the actions (as well as the intentions) of other humans or robots \cite{adams2005human, steinfeld2006common}. 
\begin{figure}[ht!]
\centering
\includegraphics[width=0.45\textwidth]{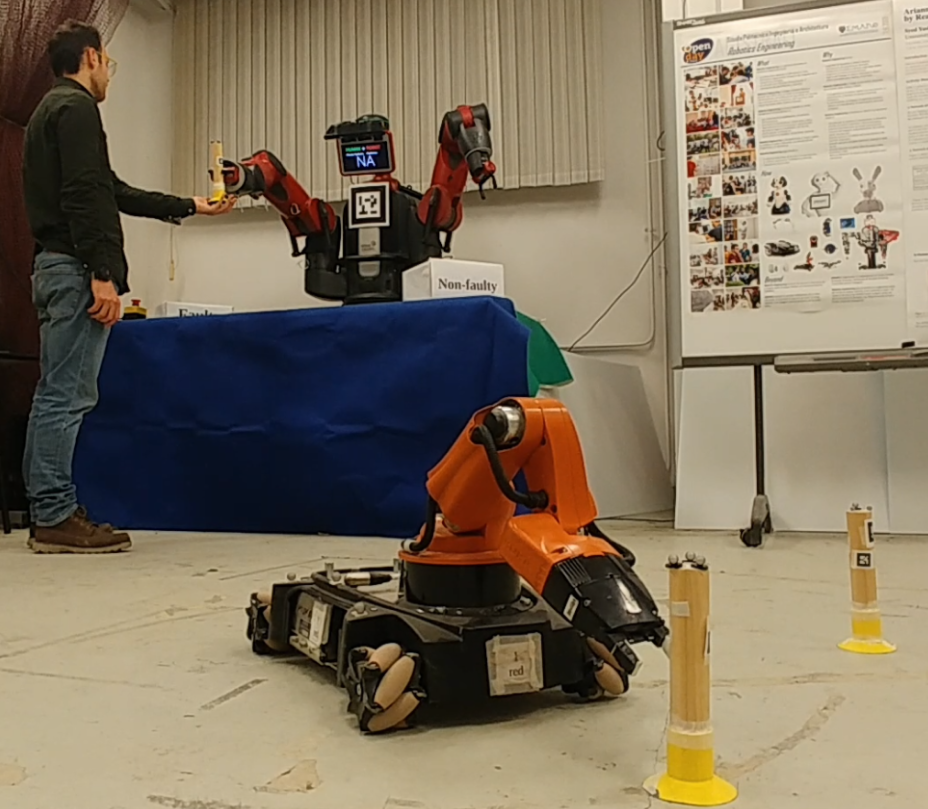}
\caption{A human operator and and two robots collaborating in a product defect inspection scenario: the mobile manipulator supplies a human operator and the dual-arm manipulator with objects to inspect.}
\label{fig:scene}
\end{figure}
The main goal of this paper is to extend the human-robot collaboration model proposed in \cite{darvish2018interleaved}, referred to as \textsc{FlexHRC}, along two directions.
On the one hand, to allow for a collaboration model taking multiple, heterogeneous robots into account, while the original work in \cite{darvish2018interleaved} considered models with one human operator and one robot.
On the other hand, introduce a use case whereby a human operator and a robot must collaboratively perform a defects inspection, whereas the original work focused on assembly tasks.

The scenario we consider is shown in Figure \ref{fig:scene}.
A mobile manipulator (in our case, a Kuka youBot) picks-up objects to be inspected (wooden pieces) from a warehouse area (a marked region in the workspace), and carries them out to deliver them to human operators or another robot (in our case, a dual-arm Baxter manipulator) for inspection \cite{Kattepur2019}.
When the object to inspect is delivered to human operators, these undertake the foreman task \cite{Huber2010assist, Glasauer2010interacting}, and then the object is passed to the manipulator for a further vision-based inspection.
Afterwards, the manipulator sorts the object out as \textit{faulty} or \textit{non faulty} in two different boxes. 
Scenarios modelling defects inspection impose functional requirements which are partially in overlap with the ones considered in \cite{darvish2018interleaved} for the assembly of semi-finished products.
The main functional requirement in quality control is the \textit{validation of products quality} with a reliable estimation. 
In an HRC process, such a requirement can be met by a double-check carried out by an expert operator in case the defects classification accuracy as provided by the robot is below a pre-specified threshold.
However, differently from the work in \cite{kawaguchi1995internal, oh2009bridge, cho2013inspection}, whereby a visual inspection is carried out by a robot, in order to validate the quality of products an integration of auditory, tactile, and visual perception is likely to be needed \cite{Spence2006auditory, Garrett2001effects}.
Such an integration is still an open issue and it is not considered in this paper.

This paper introduces and discusses \textsc{ConcHRC}, a framework extending \textsc{FlexHRC} that addresses the need for concurrent, multi human-robot collaboration in industrial environments, and validates the models in an inspection use case.
The novelty of the approach is two-fold:
(i) the design and development of an AND/OR graph based multi human-robot collaboration model that allows for concurrent, modelled, operations in a team made up of multiple human operators and/or robots;
(ii) the description of a particular instance of such a cooperation model, implemented within an existing human-robot collaboration architecture, and extending it whereby a human operator, a mobile manipulator, and a dual-arm manipulator collaborate for a defect inspection purpose.
In the paper, the focus is on the concurrent HRC model for the quality control task, and therefore we decided to simplify the robot perception system.

The paper is organized as follows. 
Section \ref{sec:RelatedWork} discusses related work.
Section \ref{sec:SystemArchitecture} introduces the \textsc{ConcHRC} architecture, and Section \ref{sec:concurrent} formalises the concurrent model. 
Section \ref{sec4} lays the experimental scenario and the related discussion.
Conclusions follow. 

\section{Related Work}
\label{sec:RelatedWork}

For a natural human-robot collaboration, different aspects such as safety, robot perception, task representation, and action execution must be considered when designing a collaborative-friendly workspace \cite{goodrich2008human, lasota2017survey}.
This paper focuses on task representation when multiple human operators and/or robots group as a team to reach a common goal, which is \textit{a priori} known to all collaborators, either humans or robots. 
The uncertainties in perception, task representation and reasoning that a robot must face increase when collaborating with humans, because a natural cooperation, i.e., a one in a way similar to human-human teams \cite{meneweger2015working}, may require the robot to \textit{make sense} of or even anticipate human intentions.
The need arises to provide robots with reasoning capabilities about the state of the human-robot cooperation process, suitable to be executed online.

Although approaches based on offline planning and task allocation fulfil a requirement related to the effectiveness of the collaboration \cite{johannsmeier2016hierarchical, lemaignan2017artificial}, they neither ensure such a natural collaboration nor address its intrinsic uncertainties. 
Differently, the approaches described in \cite{hawkins2014anticipating, levine2014concurrent, darvish2018flexible, darvish2018interleaved} are aimed at enhancing the naturalness and the flexibility of the collaboration based on online task allocation and/or contingency plans, such that the robot is able to adapt to human decisions on the spot and uncertainties. 
Such flexibility requires a rich perception for recognising human actions as well as the collaboration state \cite{darvish2018interleaved}.

Some of the methods applied for robot action planning in collaboration scenarios include \textit{Markov Decision Processes} \cite{claes2014human, crandall2018cooperating}, \textit{Task Networks} \cite{levine2014concurrent, lemaignan2017artificial}, \textit{AND/OR graphs} \cite{xie2010dynamic, johannsmeier2016hierarchical, darvish2018flexible}, and STRIPS-based \textit{planners} \cite{capitanelli2018manipulation}.
Among these methods, finding the priors and the reward function for Markov Decision Processes and the exponential growth of the computational load of STRIPS-based planners make them very difficult to be adopted in practice. 
Task Networks and AND/OR graphs ensure that the generated collaboration models are in accordance with domain expert desiderata, hence guaranteeing shared \textit{mental} models between human operators and robots.
In order to allocate tasks to human operators or robots, and to meet such collaboration constraints as limited resources, a common approach in the literature is to maximise the overall utility value of the collaboration \cite{tsarouchi2017human} on the basis of multi-objective optimisation criteria.
However, in these examples the number of human operators or robots is limited.

In order to enhance the efficiency of the collaboration, and to face the inherent limitations owing to workspace constraints, human skills, and robot capabilities, an approach can be to raise the number of human operators or heterogeneous robots involved in the collaboration.
To this aim, human operators and robots must schedule their actions according to resources, timings, and skill constraints. 
An example can be found in \cite{toussaint2016relational} whereby concurrent cooperation models are formalised according to relational activity processes. 
The authors in that study model the cooperation and predict future actions using a Monte Carlo method along with learning by demonstration.
A similar approach is adopted in \cite{smith1999temporal}, whereby a temporal graph plan with the consideration of action durations has been applied. 
Another illustration of concurrent HRC, with a probabilistic formulation due to uncertainties, is presented in \cite{weld2008planning}, where a concurrent Markov Decision Process is adopted.

In previous work \cite{darvish2018flexible, darvish2018interleaved} where we demonstrated a flexible collaboration between human operators and robots, this paper extends the notion of AND/OR graph to a concurrent model, and adopts it to model multi human-robot collaboration scenarios. 
This is further detailed in Section \ref{sec:concurrent}.

\section{System's Architecture}
\label{sec:SystemArchitecture}

\begin{figure*}[t!]
\centering
\includegraphics[width=0.73\textwidth]{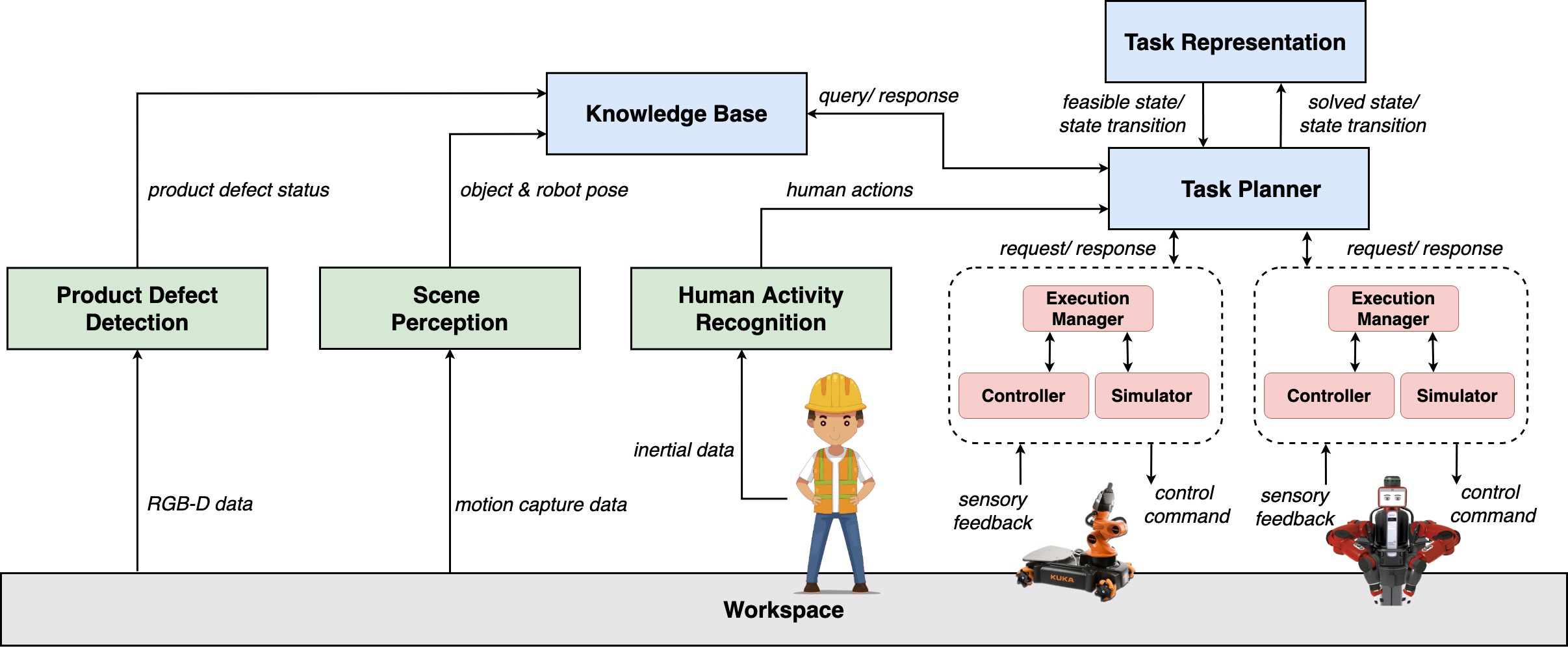}
\caption{System's architecture for a multi human-robot collaboration model in a defects detection scenario.}
\label{fig:arch}
\end{figure*}

Figure \ref{fig:arch} depicts the overall architecture of the \textsc{ConcHRC} framework.
The architecture is made up of three layers, including a \textit{perception} layer in green, a \textit{representation} layer in blue, and an \textit{action} layer in red. 
The perception layer provides information regarding the activities carried out by human operators, a part's defect status, and object locations in the robot workspace.
The representation layer forms the concurrency model, stores the necessary knowledge, and manages task execution to reach the collaboration goal.
The action level simulates and executes robot actions.

The perception layer encapsulates three modules, which are called \textit{Human Activity Recognition}, \textit{Product Defect Detection}, and \textit{Scene Perception}.
The latter two modules provide the \textit{Knowledge Base} module with information about the status of the workspace, human operators, and robots, whereas the former communicates detected human activities to the \textit{Task Planner}. 
\textit{Human Activity Recogntion} obtains inertial data originating from wearable sensors worn by human operators, and run a series of algorithms to detect and classify performed actions.
Those are modelled using \textit{Gaussian Mixture Modelling} (GMM) and \textit{Regression} \cite{bruno2014using, darvish2018flexible}.
In our setup, defects detection is considered as a classification problem. \textit{Product Defect Detection} exploits the images coming from a robot-centric camera to detect defects.


The action layer is made up of three modules, namely \textit{Robot Execution Manager}, \textit{Simulator}, and \textit{Controller}.
The \textit{Robot Execution Manager} module receives discrete, symbolic commands from the \textit{Task Planner}, maps them to actual values, and drives the behaviour of the \textit{Controller} or the \textit{Simulator}.
This module retrieves information about the workspace, human operators and robots from the \textit{Knowledge Base}.
The \textit{Robot Execution Manager} is in charge of sequencing robot behaviours, on the basis of the plan as provided by the \textit{Task Representation} module. 
It also provides an acknowledgement to the \textit{Task Planner} upon the execution of a command by the robots.
The \textit{Simulator} module is aimed at predicting the outcome of robot behaviours before their actual execution. 
It simulates a closed-loop model of the robot and the controller, by solving the ordinary differential equations online.
The \textit{Controller} receives the configuration (in joint space) or the task space command (in the Cartesian space) from the \textit{Robot Execution Manager}. 
It computes the joints velocity reference values at each control time step to the robot, while receiving feedback from it \cite{Simetti2015}. 



The representation layer embeds \textit{Task Representation}, \textit{Task Planner}, and the \textit{Knowledge Base} module. 
In the \textsc{ConcHRC}, an AND/OR graph with several layers represents the collaborative task \cite{darvish2018flexible}. 
In order to model concurrency in a multi-agent collaboration scenario, the AND/OR graph based \textsc{FlexHRC} framework has been extended, as described in the next Section.  
Along with the AND/OR graph, the \textit{Task Planner} module is in charge of decision making and the adaptation of the ongoing parallel tasks.
To do so, the \textit{Task Planner} provides a set of achieved cooperation states or transitions between states to the \textit{Task Representation} module, and receives the set of allowed cooperation states and transitions with the associated costs to follow. 
Later, it associates each state or state transition with an ordered set of actions, and according to the workspace's, human operator's, and robot's status, along with  online simulation results, it assigns actions to the either human operators or robots.
Finally, it informs each human operator or robot involved in the cooperation about the action to follow. 
Once an action is carried out, it receives the acknowledgement from the action level and updates its internal structure.
The \textit{Knowledge Base} stores all relevant information to make the cooperation progress, as better described in \cite{darvish2018flexible}.

\section{A Concurrent Model for Multi-agent Cooperation}
\label{sec:concurrent}

In this Section, we describe first a multi human-robot cooperation model based on a $1$-layer AND/OR graph, then we consider an extended $n$-layer AND/OR graph, and finally a concurrent model based on a constrained $n$-layer configuration, which we refer to as a $c$-layer AND/OR graph.

\subsection{$1$-layer AND/OR Graphs}
\label{sec:1-layer}

In order to formalise the multi human-robot cooperation process in \textsc{ConcHRC} we adopt AND/OR graphs \cite{de1990and, luger2009artificial, russell2010artificial}, as discussed above.
An AND/OR graph allows for representing \textit{procedures} to follow, which can be decomposed in subproblems as parts of the graph, as well as the logic \textit{relationships} among them, i.e., the graph interconnectivity.
The root node conventionally represents the goal state of the process being modelled, and achieving the goal means traversing the graph from leaf nodes to the root node via intermediate nodes and hyper-arcs according to its structure.

A $1$-layer AND/OR graph $G$ can be formally defined as a $2$-ple $\langle N, H \rangle$ where $N$ is a set of $|N|$ nodes, and $H$ is a set of $|H|$ hyper-arcs. 
An hyper-arc $h \in H$ induces the set $N_c(h) \subset N$ of its \textit{child} nodes, and the singleton $N_p(h) \subset N$ made up of a \textit{parent} node, such that
\begin{equation}
h: N_c(h) \rightarrow N_p(h).
\label{eq:hyper_arc_trans_simple}
\end{equation}
Furthermore, we define $n \in N$ as a \textit{leaf} node if $n$ is not a parent node for any hyper-arc, i.e., if $h \in H$ does not exist such that $n \in N_p(h)$, or as a \textit{root} node if it is the only node that is not a child node for any hyper-arc, i.e., if $h \in H$ does not exist such that $n \in N_c(h)$.

In a multi human-robot cooperation scenario, each node $n \in N$ represents a cooperation \textit{state},
e.g., \textit{faulty object inside box}, whereas each hyper-arc $h \in H$ represents a (possibly) \textit{many-to-one} transition among states, i.e., activities performed by human operators and/or robots, which make the cooperation move forward, such as \textit{the robot puts the faulty object into the box}.
The relation among child nodes in hyper-arcs is the logical \textit{and}, whereas the relation between different hyper-arcs inducing on the same parent node is the logical \textit{or}, i.e., different hyper-arcs inducing on the same parent node represent alternative ways for a cooperation process to move on.
Each hyper-arc $h \in H$ implements the transition in (\ref{eq:hyper_arc_trans_simple}) by checking the \textit{requirements} defined by nodes in $N_c(h)$, executing \textit{actions} associated with $h$, and generating \textit{effects} compatible with the parent node.
Each hyper-arc $h \in H$ executes an ordered set $A(h)$ of \textit{actions}, such that
\begin{equation}
A(h) = (a_1, \ldots,a_{|A|}; \preceq),
\end{equation}
where the precedence operator $\preceq$ defines the pairwise expected order of action execution.
The sequence can be scripted or planned online \cite{capitanelli2018manipulation}.
Before an hyper-arc $h$ is executed, all actions $a \in A(h)$ are marked as \textit{undone}, i.e., ${\textsf{done}(a) \leftarrow false}$.
When one action $a$ is executed by any agent, its status changes to ${\textsf{done}(a) \leftarrow true}$.
An hyper-arc $h \in H$ is marked as \textit{solved}, i.e., ${\textsf{solved}(h) \leftarrow true}$ \textit{iff} all actions $a \in A(h)$ are done in the expected order.
In a similar way, nodes $n \in N$ may be associated with a (possibly ordered) set of \textit{processes} $P(n)$, which are typically \textit{robot} behaviours activated in a cooperation state but not leading to a state transition.

It is possible to introduce the notion of feasibility.
A node $n \in N$ is \textit{feasible}, i.e., $\textsf{feasible}(n) \leftarrow true$, \textit{iff} a solved hyper-arc $h \in H$ exists, for which $n \in N_p(h)$, and $\textsf{met}(n) \leftarrow false$, i.e.,  
\begin{equation}
\exists h \in H. \left(\textsf{solved}(h) \cap n \in N_p(h) \cap \neg \textsf{met}(n)\right).
\label{eq:feasible_node}
\end{equation}
All leaf nodes in an AND/OR graph are usually feasible at the beginning of the multi human-robot cooperation process, which means that the cooperation can be performed in many ways.
An hyper-arc $h \in H$ is \textit{feasible}, i.e., $\textsf{feasible}(h) \leftarrow true$, \textit{iff} for each node $n \in N_c(h)$, $\textsf{met}(n) \leftarrow true$ and $\textsf{solved}(h) \leftarrow false$, i.e.,
\begin{equation}
\forall n \in N_c(h).\left(\textsf{met}(n) \cap \neg \textsf{solved}(h)\right).
\label{eq:feasible_hyper_arc}
\end{equation}
Once an hyper-arc $h_i \in H$ is solved, all other feasible hyper-arcs $h_j \in H\setminus\{h_i\}$, which share with $h_i$ at least one child node, i.e., $N_c(h_i) \cap N_c(h_j) \neq \emptyset$, are marked as unfeasible, in order to prevent the cooperation process to consider alternative ways to cooperation that have become irrelevant.

Given and AND/OR graph, the multi human-robot cooperation process is modelled as a \textit{graph traversal} procedure which, starting from a set of leaf nodes, must reach the root node by selecting hyper-arcs and reaching states in one of the available \textit{cooperation paths}, depending on the feasibility statuses of nodes and hyper-arcs.
According to the graph structure,
multiple cooperation paths may exist, meaning that multiple ways to solve the task may be equally legitimate. 
The traversal procedure dynamically follows the cooperation path that at any time is characterised by the lowest cost.
The entire algorithm has been described in \cite{darvish2018flexible,darvish2018interleaved}.
The traversal procedure suggests to human operators agents actions in the hyper-arcs that are part of the path, and sends to robots the actions they must execute.
Human operators can override the suggestions at any time, executing different actions, which may cause the graph to reach a state not part of the current path.
When this happens, \textsc{ConcHRC} tries to progress from that state onwards \cite{darvish2018flexible,darvish2018interleaved}.
This mechanism enables \textsc{ConcHRC} to pursue an optimal path leading to the solution, while it allows human operators to choose alternative paths.
As long as the multi human-robot cooperation process unfolds, and the AND/OR graph is traversed, we refer with $N_f$ and $H_f$ to the sets of \textit{currently} feasible nodes and hyper-arcs, respectively.
We say that an AND/OR graph $G$ is \textit{solved}, i.e., $\textsf{solved}(G) \leftarrow true$, \textit{iff} its root node $r \in N$ is met, i.e., $\textsf{met}(r) \leftarrow true$.
Otherwise, if the condition $N_f \cup H_f = \emptyset$, i.e., there are no feasible nodes nor hyper-arcs, then the multi human-robot cooperation process fails, because there is no feasible cooperation path leading to the root node.

\subsection{$n$-layer AND/OR Graphs}
\label{sec:n-layer}

A $n$-layer AND/OR graph $G^n$ can be recursively defined as a $2$-ple $\langle \Gamma, \Theta \rangle$ where $\Gamma$ is an ordered set of $|\Gamma|$ \textit{up to} $(n-1)$-layer AND/OR graphs, such that:
\begin{equation}
\Gamma = \left(G_1, \ldots, G_{|\Gamma|}; \preceq \right),
\label{eq:gamma_set}
\end{equation}
and $\Theta$ is a set of $|\Theta|$ pairwise transitions between them.
In (\ref{eq:gamma_set}), the AND/OR graphs are ordered according to their layer.
Lower-layer AND/OR graphs are characterised by a decreasing level of abstraction, i.e., they are aimed at modelling the HRC process more accurately.
Transitions in $\Theta$ define how different AND/OR graphs in $\Gamma$ are connected, and in particular model the relationship between graphs belonging to different layers. 

If we recall (\ref{eq:hyper_arc_trans_simple}) and we contextualise it for an AND/OR graph $G^n = \langle N^n, H^n \rangle$, we observe that a given hyper-arc in $H^n$ represents a mapping between the set of its child nodes and the singleton parent node.  
We can think of a generalised version of such a mapping to encompass a whole AND/OR graph $G^{n-1} = \langle N^{n-1}, H^{n-1} \rangle$, where the set of child nodes is constituted by the set $N^{n-1}_L$ of leaf nodes, and the singleton parent node by the graph's root node $r^{n-1} \in N^{n-1}$.
As a consequence, a transition $T \in \Theta$ can be defined between a hyper-arc $h \in H^n$ and an entire AND/OR graph $G^{n-1}$, such that
\begin{equation}
T: h \rightarrow G^{n-1}, 
\label{eq:graph_transition}
\end{equation}
subject to the fact that appropriate mappings can be defined between the set of child nodes of $h$ and the set of leaf nodes of the deeper graph, i.e.,
\begin{equation}
M_1: N_c(h) \rightarrow N_L \in N^{n-1}, 
\label{eq:mapping_1}
\end{equation}
and between the singleton set of parent nodes of $h^n$ and the root node of the deeper graph, i.e.,
\begin{equation}
M_2: N_p(h) \rightarrow r^{n-1} \in N^{n-1}. 
\label{eq:mapping_1}
\end{equation}
Mappings $M_1$ and $M_2$ must be such that the corresponding information in different layers should be \textit{semantically equivalent}, i.e., it should represent the same information with a different representation granularity. 
The same applies for $N_p(h)$ and the root of $G^{n-1}$.
Once these mappings are defined, it easy to see that $G^n$ has a tree-like structure, where graphs in $\Gamma$ are nodes and transitions in $\Theta$ are edges.

An AND/OR graph $G^n$ is feasible, i.e., $\textsf{feasible}(G^n) \leftarrow true$ \textit{iff} it has at least one feasible node or hyper-arc.
If a transition $T \in \Theta$ exists in the form (\ref{eq:graph_transition}), a hyper-arc $h \in H^n$ is feasible \textit{iff} the associated AND/OR graph $G^{n-1}$, is feasible, i.e., 
\begin{equation}
\forall T.\left(\textsf{feasible}(h) \leftrightarrow \textsf{feasible}(G^{n-1})\right).
\label{eq:feasibility_hierarchical_graph}
\end{equation}
As a consequence, when the nodes in $N^{n-1}_L$ of $G^{n-1}$ becomes feasible, the hyper-arc $h$ in $G^n$ becomes feasible as well.
Furthermore, the hyper-arc $h$ is solved \textit{iff} the associated AND/OR graph $G^{n-1}$ is solved, i.e.,
\begin{equation}
\forall T.\left(\textsf{solved}(h) \leftrightarrow \textsf{solved}(G^{n-1})\right).
\label{eq:solve_hierarchical_graph}
\end{equation}

\subsection{$c$-layer AND/OR Graphs}
\label{sec:n-layer}

A concurrent AND/OR graph is modelled as a restriction of a $n$-layer AND/OR graph whereby the $n$-th layer is aimed at modelling the termination condition for the whole hierarchy of $(n-1)$-layer graphs, and the latter model different, concurrent activities part of the HRC process. 
A $c$-layer AND/OR graph must also specify if and how nodes belonging to separate lower-layer graphs are synchronised.

Analogously to an $n$-layer graph, a $c$-layer AND/OR graph $G^c$ can be defined as a $2$-ple $\langle \Gamma^c, \Theta^c \rangle$ where $\Gamma^c$ is an ordered set of $|\Gamma^c|$ \textit{up to} $(n-1)$-layer AND/OR graphs, such that:
\begin{equation}
\Gamma^c = \left(G_1, \ldots, G_{|\Gamma^c|}; \preceq \right),
\label{eq:concurrent_set}
\end{equation}
and $\Theta^c$ is a set of $|\Theta^c|$ pairwise transitions between them.

Whilst the considerations related to $n$-layer AND/OR graphs apply for $c$-layer AND/OR graphs, the composition of the constituting sets of nodes and hyper-arcs may differ.
Let us recall that for a generic AND/OR graph $G$ we refer to $N$ as its set of nodes, and with $H$ as its set of hyper-arcs, and let us consider two AND/OR graphs $G_i$ and $G_j \in \Gamma^c$.
Let us limit ourselves to a weak notion of independence between graphs.
We consider $G_i$ and $G_j$ as mutually independent \textit{iff} there is no node in $G_i$ (respectively, $G_j$) that needs to be met before another node of $G_j$ (respectively, $G_i$).
If this is the case, $G_i$ (respectively, $G_j$) can be modelled as a generic $n$-layer AND/OR graphs $\langle N_i, H_i \rangle$ (respectively, $\langle N_j, H_j \rangle$). 
Otherwise, if $G_i$ is dependent on $G_j$, i.e., a node $n_j$ in $G_j$ must be met before another node $n_i$ in $G_i$ can be \textsf{met}, we need to formally model it as an external dependence.

To this aim, and in general terms, we augment the set of nodes $G_i$ with a set of dependence nodes, whose associated logic predicates \textsf{met} are entangled with the corresponding nodes in $G_j$, such that their truth values always correspond.
A node $n^e$ of an AND/OR graph $G_i$ is said to be \textit{entangled} with a node $n_j$ of an AND/OR graph $G_j$, with $i \neq j$, \textit{iff} for that node
\begin{equation}
\textsf{met}(n^e) \leftrightarrow \textsf{met}(n_j)
\label{eq:entangled_node}
\end{equation}
and $n^e$ is a leaf node for $G_i$, i.e., $N_c(n^e) = \emptyset$.
Then, a \textit{dependent} AND/OR graph $G_i$ is defined as a $2$-ple $\langle N^c_i, H^c_i \rangle$, such that 
$N^c_i = N_i \cup \{n^e_1, \ldots, n^e_\eta\}$, i.e., the union between the set of nodes $N_i$ as if the graph were not dependent on any other graph, plus the set of the entangled nodes, and  
$H^c_i = H_i \cup \{h^e_1, \ldots, h^e_\lambda\}$, i.e., the union between the set of hyper-arcs $H_i$ as if the graph were not dependent on any other graph, plus the set of the hyper-arcs reliant on entangled nodes.


\section{Experimental Validation}
\label{sec4}

\subsection{Implementation of the Multi Human-Robot Collaboration Process for Defects Inspection}

In order to validate the effectiveness of \textsc{ConcHRC}, we implemented an abstract defects inspection scenario. 
The scenario has been briefly described in the Introduction, and is represented in Figure \ref{fig:scene}. 
A Kuka youBot omni-directional mobile manipulator is used to pick up objects from a \textit{warehouse area}, and brings them close to the \textit{defects inspection} cell, where a human operator and a dual-arm Baxter robot are expected to collaborate.
The youBot and the objects to be manipulated are localised in the workspace using an external motion capture system based on passive markers, i.e., a system composed of $8$ OptiTrack-Flex $13$ motion capture cameras.
Baxter is provided with the standard grippers, and is equipped also with a RGB-D camera mounted on the robot \textit{head} and pointing downward, which is used to acquire images for defects inspection.
Since, in our case, the focus is on the multi human-robot collaboration process, we decided to over-simplify the inspection, which is surrogated using QR tags corresponding to \textit{faulty}, \textit{non faulty}, \textit{Na}, respectively.
Actions carried out by human operators are perceived via their inertial blueprint via an LG G Watch R (W110) smartwatch, worn at the right wrist.
Data are transmitted through a standard WiFi link to a workstation. 
The workstation is equipped with an Intel(R) core i7-8700 @ 3.2 GHz $\times$ 12 CPUs and 16 GB of RAM. 
The architecture is developed using C++ and Python under ROS Kinetic.

There are upper bounds to the maximum angular velocity of arm joints for both the Baxter and the youBot, i.e., $0.6$ $rad/s$. 
Limits on the youBot's linear and angular velocities are $0.4$ $m/s$ and $0.3$ $rad/s$, respectively. 
These limits are applied to both simulated and real robots.
Action models foreseen for human activity recognition are simply \textit{pick up} and \textit{put down}.
Instead, actions used for Baxter arms include \textit{approach}, \textit{grasp}, \textit{ungrasp}, \textit{hold on}, \textit{stop}, \textit{check object status}, whereas for the youBot arm we considered only \textit{approach}.

\begin{figure}[t!]
\centering
\includegraphics[scale=0.31]{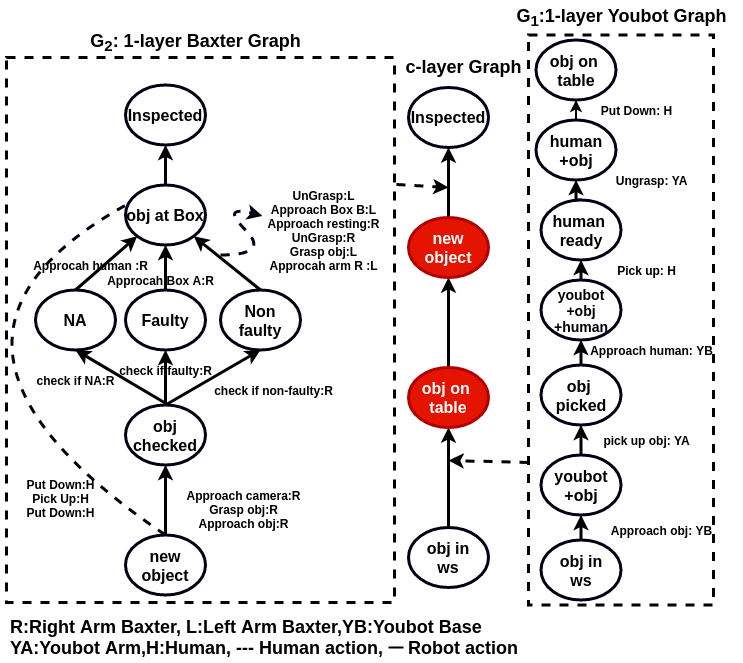}
\caption{The collaboration graph for defects inspection.}
\label{fig:andorbaxter}
\end{figure}
\begin{figure}[t!]
\centering
\includegraphics[scale=0.12]{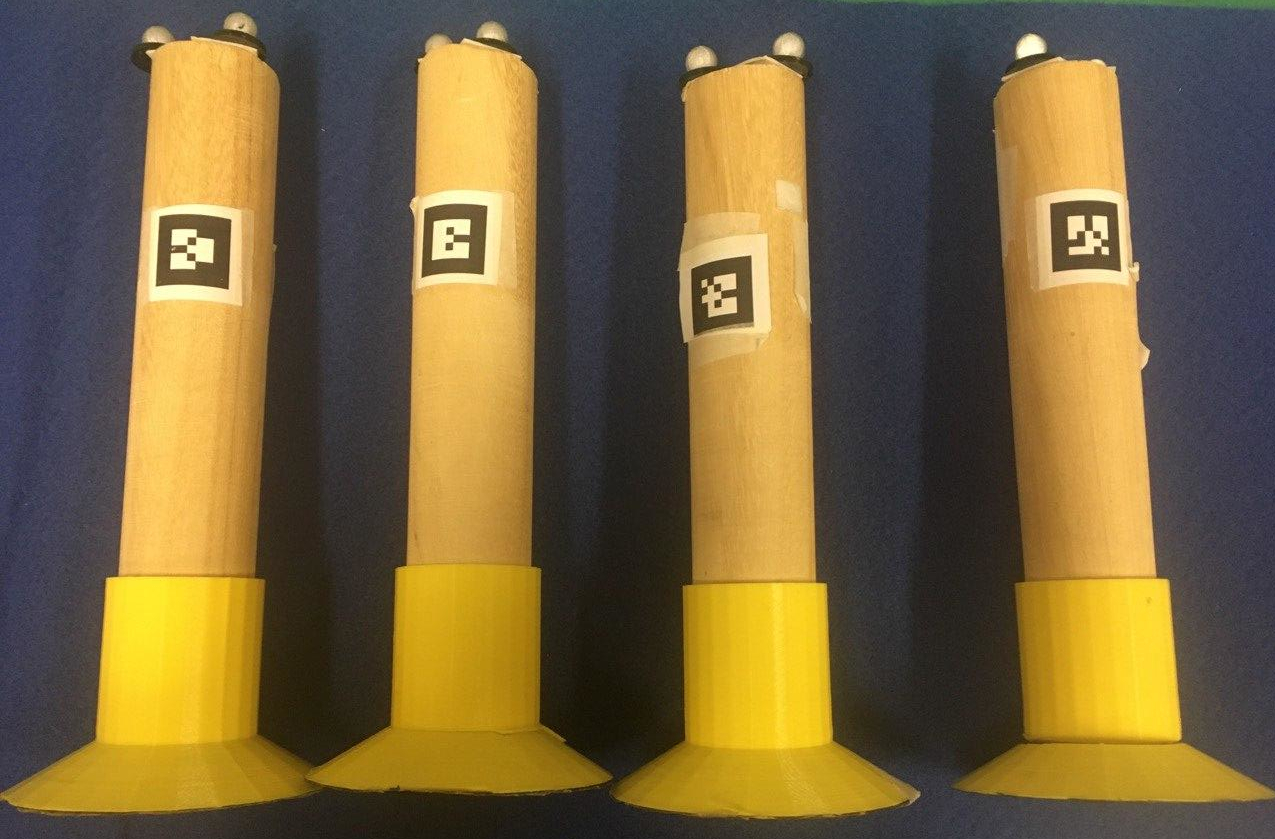}
\caption{Four tagged cylinders used in our scenario.}
\label{fig:obj}
\end{figure}

Our scenario includes three physical agents, i.e., a human operator, Baxter and youBot, but five \textit{logical} agents, i.e., the operator, the Baxter left arm, the Baxter right arm, the youBot base, and the youBot arm.
However, one planner manages both Baxter arms, and likewise one planner manages the youBot base and arm, so they are used sequentially. 
In the scenario, objects are randomly placed in the warehouse area.
Objects are cylinders labeled with three different QR code types (Figure \ref{fig:obj}).
The youBot must find each object, move towards it, pick it, take it to the area where the human operator and the Baxter are located, and hand it over the operator. 
This sequence is repeated until all objects are delivered. 
On the other side of the collaboration scenario, the Baxter starts its operations when the human operator puts down an object on the table in front of the robot.
By default, its right arm is used to pick the object up, and to check whether it is faulty, non-faulty or the defect cannot be assessed. 
If the object is faulty, it is placed in a \textit{faulty} box close to the right arm, or in case of a non-faulty object, the object is handed over to the left arm to be placed in a \textit{non-faulty} box. 
If the object level of defects cannot be assessed, then it is handed back to the human operator for an \textit{ad hoc} assessment.
This process is repeated for all objects. 

\subsection{Description of the Experiment}
 
Figure \ref{fig:andorbaxter} shows a $c$-layer concurrent AND/OR graph, which is composed of two $1$-layer AND/OR graphs, for the youBot ($G_1$) and the Baxter ($G_2$), respectively. 
Entangled nodes of both graphs are depicted in red, which makes graph $G_2$ dependent on graph $G_1$.
In order for the leaf node of $G_2$ (i.e., \textit{new object}) to be feasible, the root node of $G_1$ (i.e., \textit{obj on table}) must be met.

\begin{figure*}
\centering
\begin{subfigure}{0.24\textwidth}
  \centering
  \includegraphics[width=3.7cm]{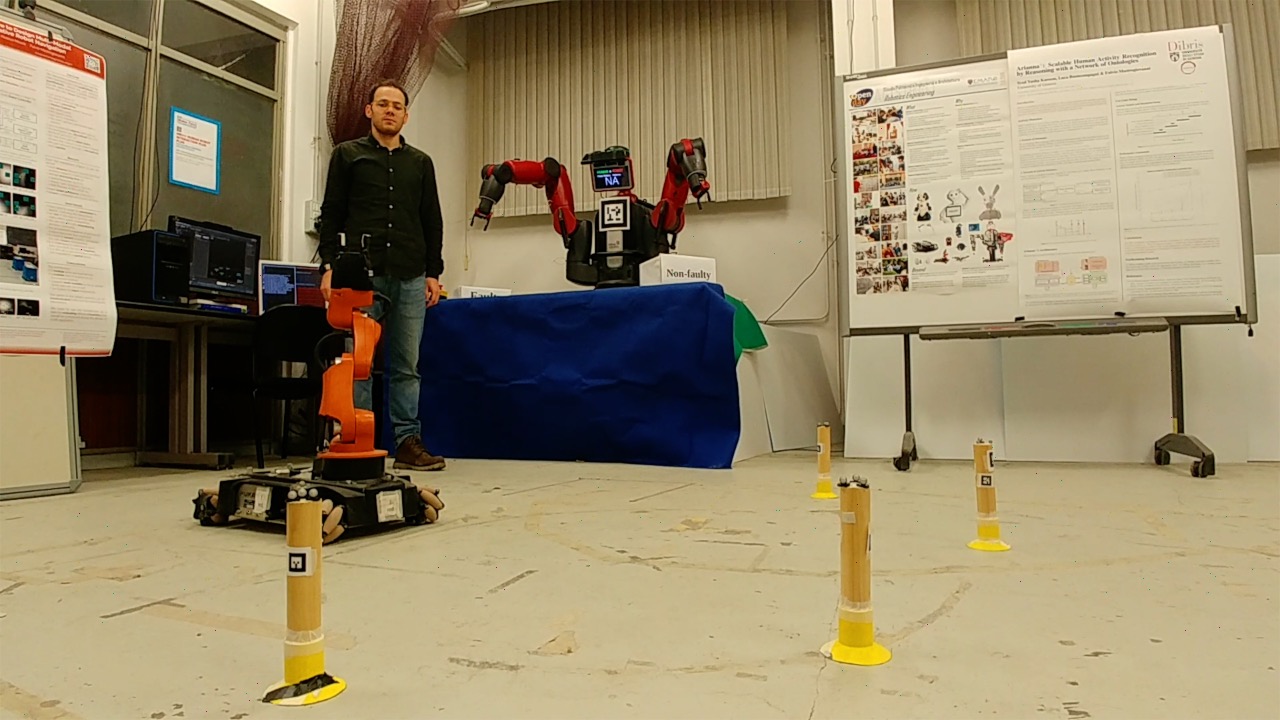}
  \caption{}
  \label{fig:sfig1}
\end{subfigure}
\begin{subfigure}{0.24\textwidth}
  \centering
  \includegraphics[width=3.7cm]{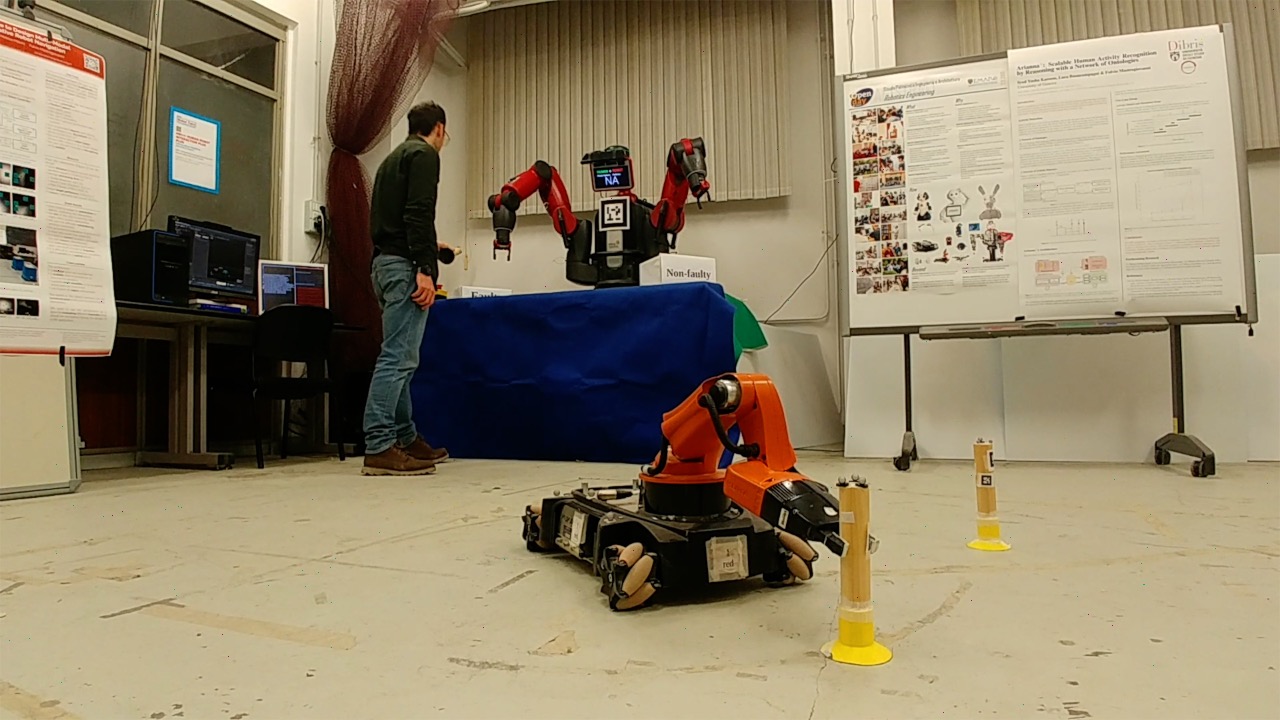}
  \caption{}
  \label{fig:sfig2}
\end{subfigure}
\begin{subfigure}{0.24\textwidth}
  \centering
  \includegraphics[width=3.7cm]{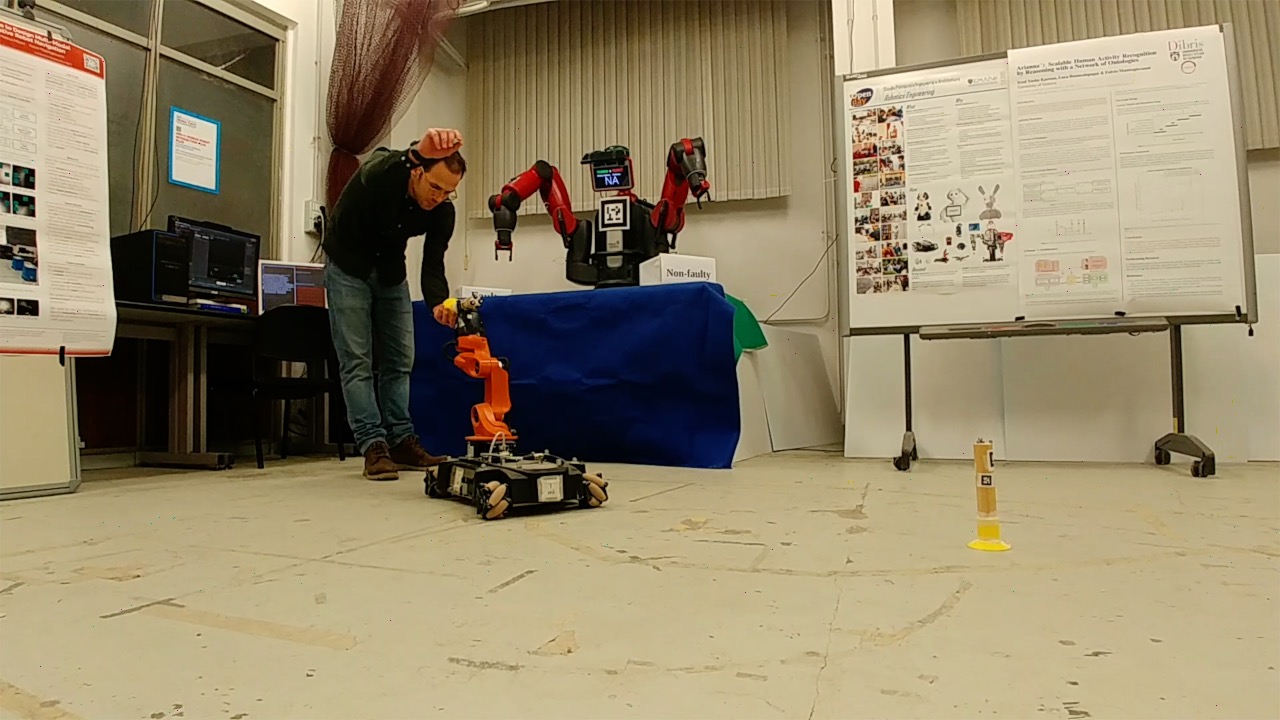}
  \caption{}
  \label{fig:sfig3}
\end{subfigure}
\begin{subfigure}{0.24\textwidth}
  \centering
  \includegraphics[width=3.7cm]{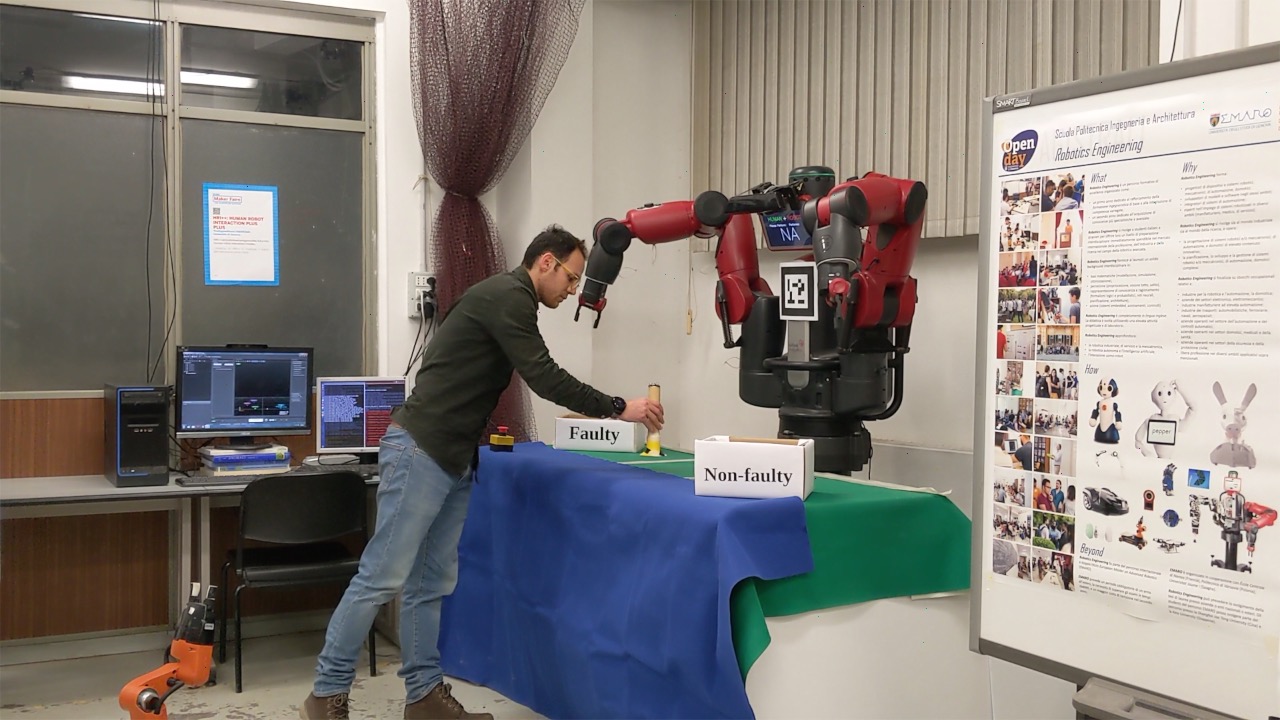}
  \caption{}
  \label{fig:sfig4}
\end{subfigure}
\begin{subfigure}{0.24\textwidth}
  \centering
  \includegraphics[width=3.7cm]{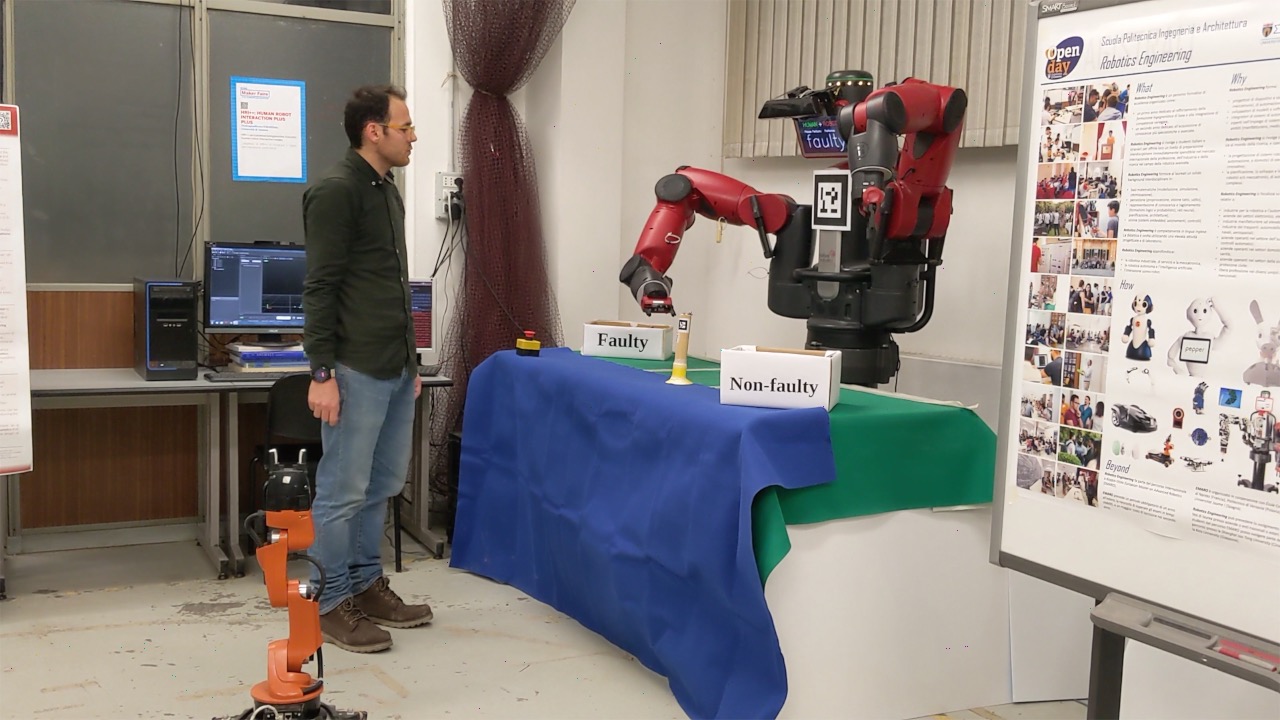}
  \caption{}
  \label{fig:sfig5}
\end{subfigure}
\begin{subfigure}{0.24\textwidth}
  \centering
  \includegraphics[width=3.7cm]{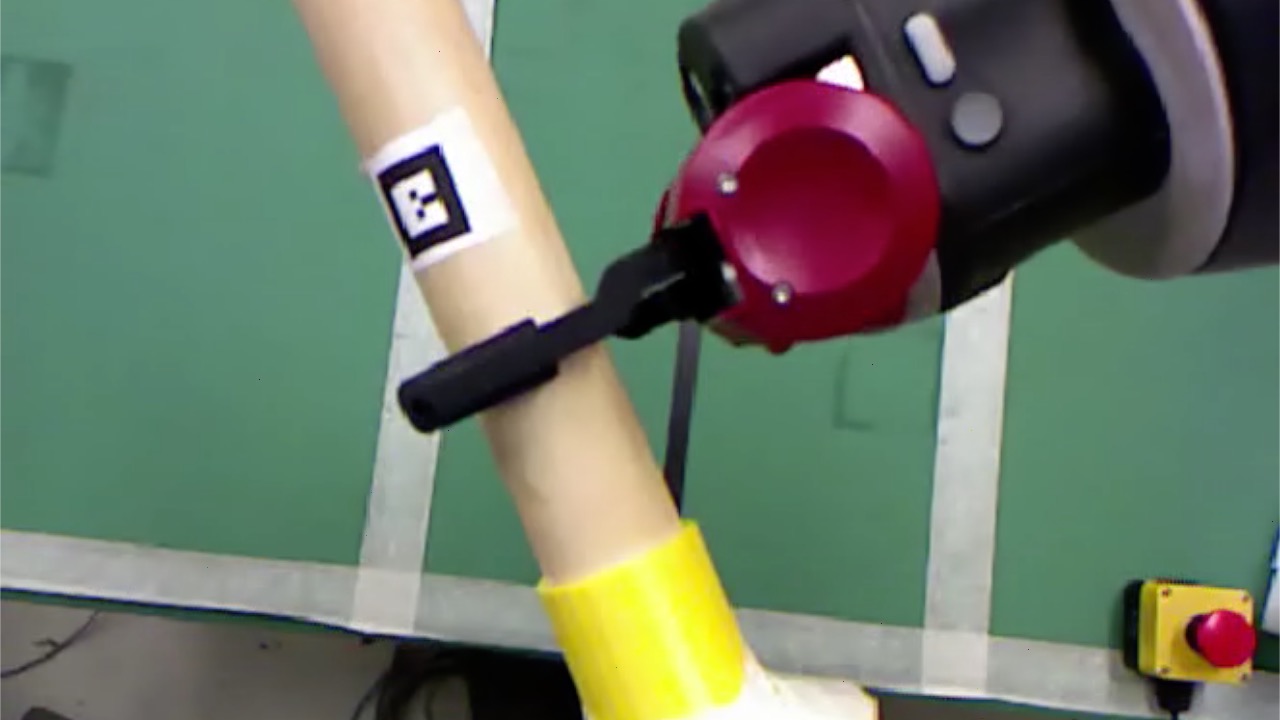}
  \caption{}
  \label{fig:sfig6}
\end{subfigure}
\begin{subfigure}{0.24\textwidth}
  \centering
  \includegraphics[width=3.7cm]{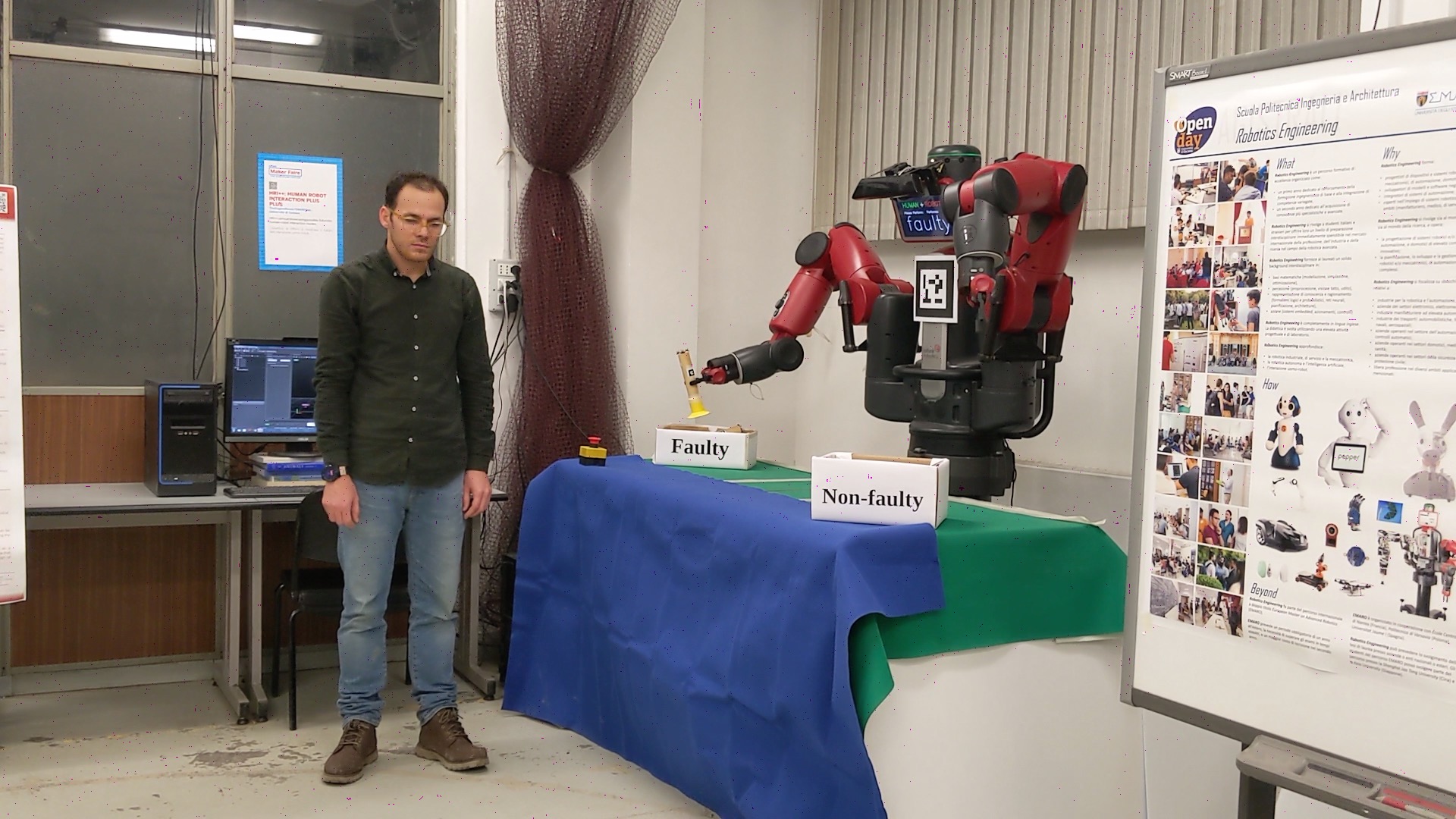}
  \caption{}
  \label{fig:sfig7}
\end{subfigure}
\begin{subfigure}{0.24\textwidth}
  \centering
  \includegraphics[width=3.7cm]{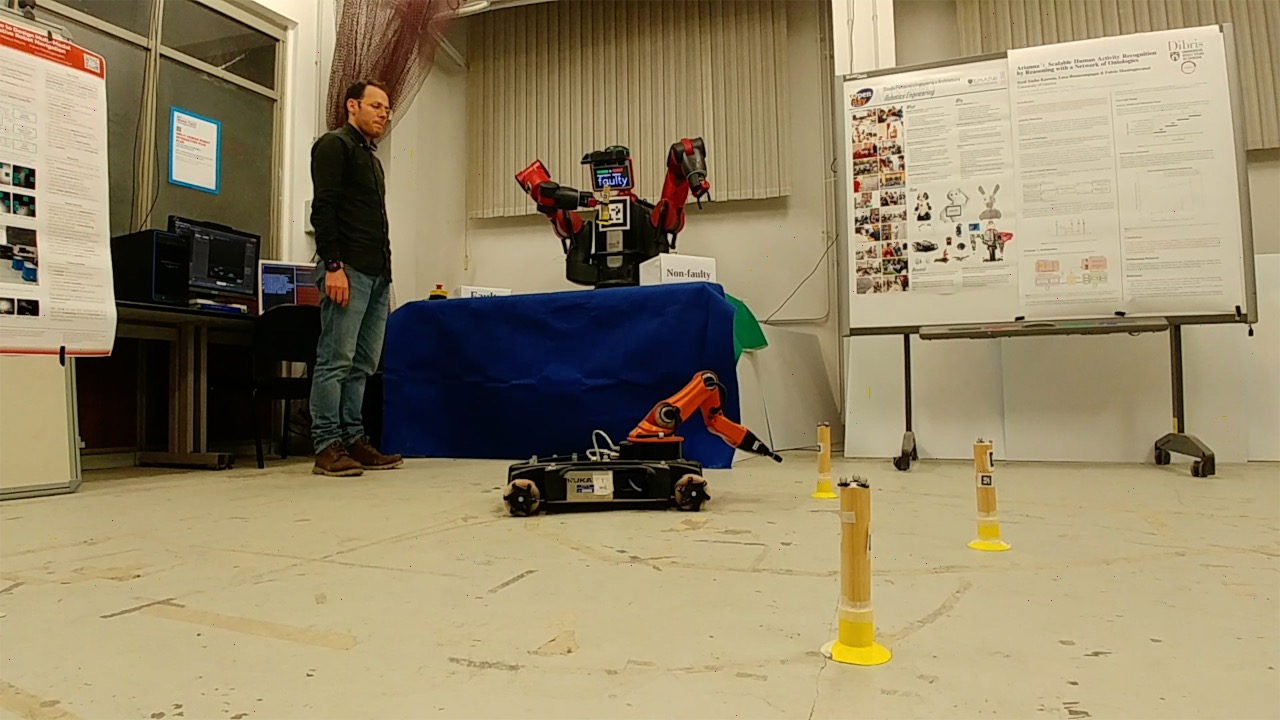}
  \caption{}
  \label{fig:sfig8}
\end{subfigure}
\begin{subfigure}{0.24\textwidth}
  \centering
  \includegraphics[width=3.7cm]{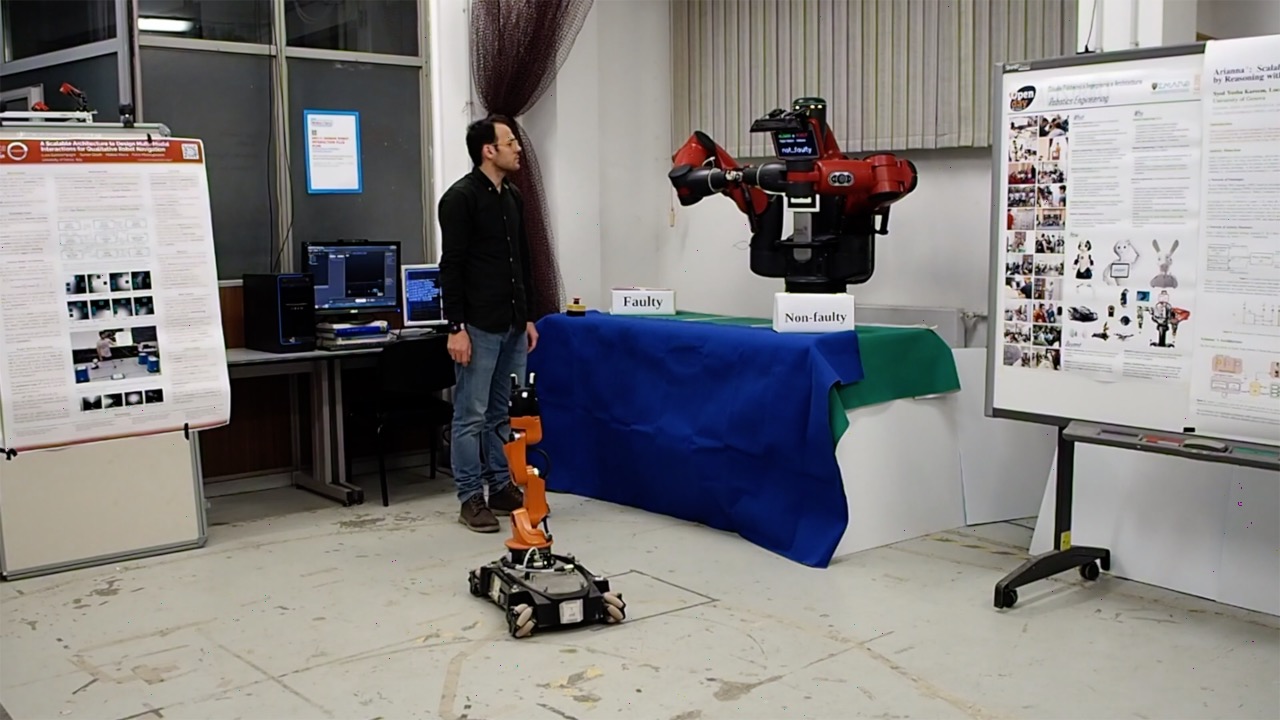}
  \caption{}
  \label{fig:sfig9}
\end{subfigure}
\begin{subfigure}{0.24\textwidth}
  \centering
  \includegraphics[width=3.7cm]{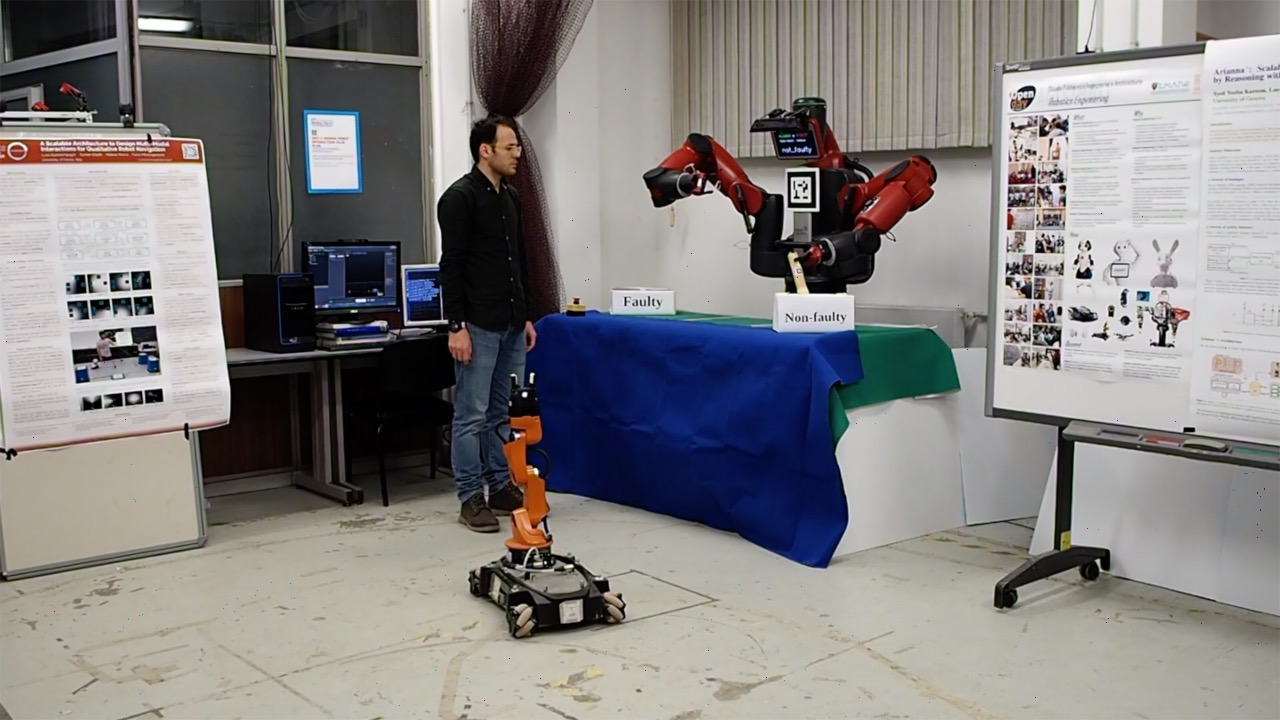}
  \caption{}
  \label{fig:sfig10}
\end{subfigure}
\begin{subfigure}{0.24\textwidth}
  \centering
  \includegraphics[width=3.7cm]{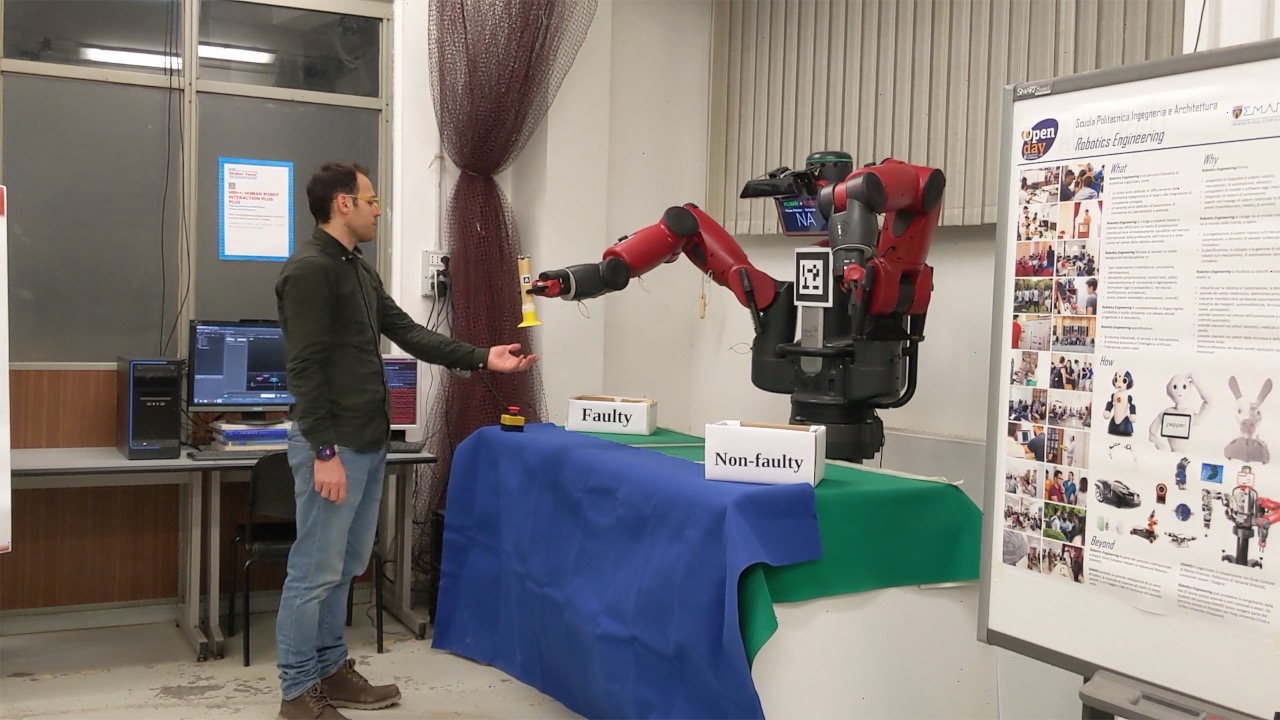}
  \caption{}
  \label{fig:sfig11}
\end{subfigure}
\begin{subfigure}{0.24\textwidth}
  \centering
  \includegraphics[width=3.7cm]{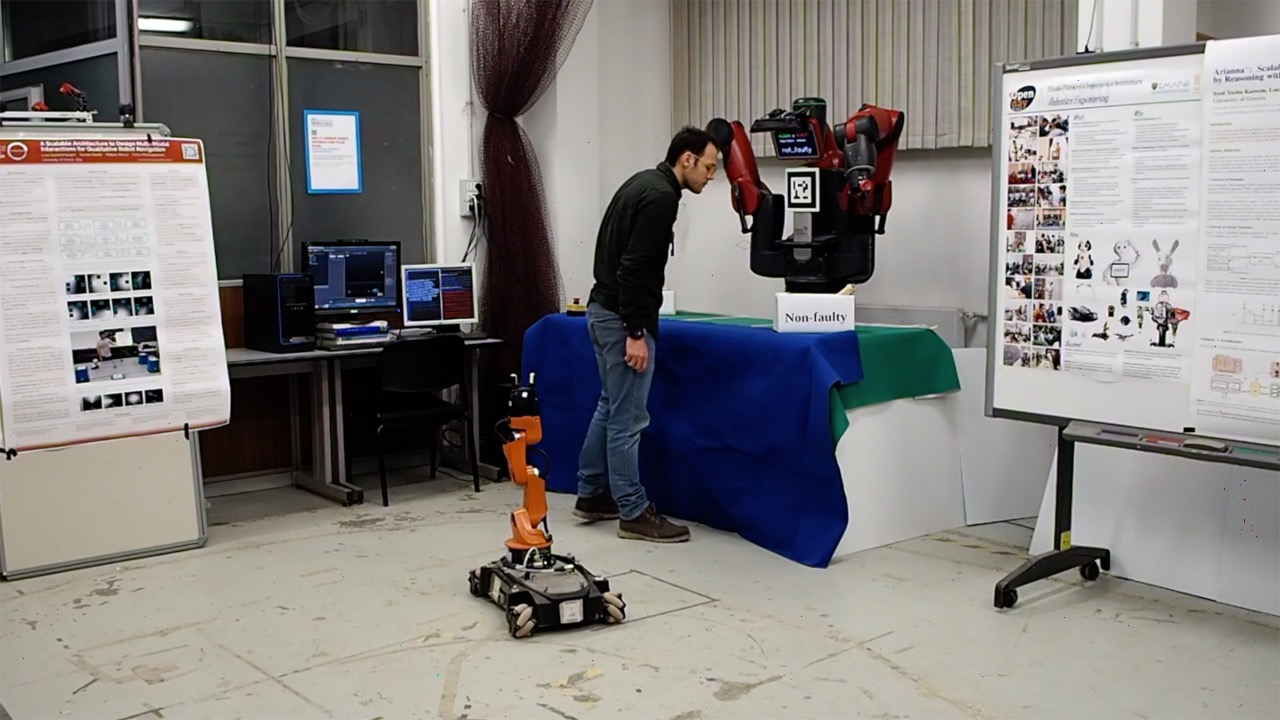}
  \caption{}
  \label{fig:sfig12}
\end{subfigure}
\caption{A typical sequence of tasks in a defects inspection experiment.}
\label{fig:exp}
\end{figure*}

During the HRC process, the human operator is typically close to the Baxter, as shown in Figure \ref{fig:exp}. 
When the youBot approaches, the operator executes a \textit{discrete gesture} moving an arm upward in order to announce a \textit{pick up} action.
Once the gesture is detected, youBot releases the object opening the end-effector to hand it over. 
Afterwards, the operator announces via a \textit{put down} gesture the fact that the object to inspect has been located on the table for the Baxter to start inspection. 

Figure \ref{fig:exp} shows a typical run of the collaboration process.
In the initial configuration, shown in Figure \ref{fig:sfig1}, both the human operator and the robots are in stand by mode.
The youBot moves towards the next object to inspect (\textit{obj in ws} state), according to graph $G_1$.
The object is selected on the basis of the time it takes to perform the whole operation in simulation.
After approching the object (\textit{youbot+obj}), the youBot's arm attempts grasping (Figure \ref{fig:sfig2}), and then picking it up (\textit{obj picked}).
In the meantime, the Baxter is waiting for human operator actions to start collaboration. 
The youBot moves towards the human operator (Figure \ref{fig:sfig3}), and  waits for a command to release the object (\textit{youbot+obj+human} state).
This is done by the operator by moving an arm upward (\textit{human ready}), which implies the youBot to open the gripper.
The operator takes the object (\textit{human+obj}) and puts it down (\textit{obj on table}) on the table.
The operator, then, can keep moving downward one arm, therefore notifying to the Baxter that an object is on the table (Figure \ref{fig:sfig4}). 
An entangled node (\textit{new object}) becomes feasible after the root node of $G_1$ is met. 
It is noteworthy that in some cases the youBot was not able to grasp objects properly, or dropped it actually before handover could occur.
Furthermore, it happened that human actions were not recognised, which required the operator to repeat them.
In these cases it is the operator's responsibility to handle the situation by taking appropriate actions in order to make the collaboration fluent.
Upon the notification of the appropriate operator gesture, the Baxter starts grasping the object (Figure \ref{fig:sfig5}) and moves it in order to place it in front of the head-mounted camera, rotating it (\textit{obj checked}) for defects inspection (Figure \ref{fig:sfig6}). 
In Figure \ref{fig:sfig7}, it is shown how the object is recognised as \textit{faulty}, and therefore the right arm places it in the \textit{faulty box} (\textit{obj at box}). 
While the Baxter is inspecting the object, the youBot continues to look for other objects (Figure \ref{fig:sfig8}). 
After a while, as shown in Figure \ref{fig:sfig9}, one of the objects is classified as \textit{non faulty}.
Since the related box cannot be reached by the Baxter right arm, an handover in-between the two arms is executed (Figure \ref{fig:sfig10}).
In case the assessment cannot be done (this is simulated with a specific QR tag), the graph reaches a \textit{NA} state, which implies that the Baxter requires the human operator to inspect the object directly (Figure \ref{fig:sfig11}).
After all objects are inspected (\textit{inspected} state), the human operator performs a check (Figure \ref{fig:sfig12}).

In order to perform a realistic computational assessment of the architecture, the whole scenario has been tested five times.
Results can be seen in Table \ref{table:1}, where times are related to the whole experiments\footnote{A video is available at https://youtu.be/0aOOeqCL2So.}. 
Statistics presented in Table \ref{table:1a} and Table \ref{table:1b} seem to indicate that the representation and planning modules together require less than 1\% of the overall execution time, whereas the major portion of collaboration time is related to human or robot actions. 
The standard deviation related to task planners and the representation modules for both robots are low enough to be neglected, and imposes no latency in collaboration proccess. 

\begin{table}[t!]
\centering
\caption{Execution times.}

\begin{subtable}[c]{0.5\textwidth}
\centering
\begin{tabular}{c c c c} 
Module              & Avg. time [s] & Avg. time $[\%]$  & Std. dev. [s] \\ 
\hline
Task Representation & 0.52          & 0.21              & 0.01 \\ 
Task Planner        & 0.02          & 0.008             & 0.003 \\
Simulator           & 3.69          & 1.49              & 0.24 \\
Baxter actions      & 203.00        & 82.00             & 5.00 \\
Human actions       & 39.00         & 15.80             & 6.00 \\  
Total               & 246.75        & 100.00            & 11.253 \\ [1ex]
\end{tabular}
\caption{Baxter-related activities.}
\label{tab:label subtable A}
\label{table:1a}
\end{subtable}

\vfill
\begin{subtable}[c]{0.5\textwidth}
\centering
\begin{tabular}{c c c c} 
Module              & Avg. time [s] & Avg. time $[\%]$  & Std. dev. [s] \\ 
\hline
Task Representation & 0.43          & 0.13              & 0.02 \\ 
Task Planner        & 0.02          & 0.00              & 0.004 \\
Simulator           & 2.74          & 0.79              & 0.40 \\
youBot actions      & 268.00        & 86.00             & 14.00 \\
Human actions       & 39.00         & 12.50             & 6.00 \\  
Total               & 310.19        & 100.00            & 20.424 \\ [1ex]
\end{tabular}
\caption{youBot-related activities.}
\label{tab:label subtable B}
\label{table:1b}
\end{subtable}

\label{tab:label all table}
\label{table:1}
\end{table}

\subsection{Discussion}

On the basis of the experiments we carried out, it is possible to make two different remarks.

The first is related to the robustness associated with the overall process.
In spite of such faults as unsuccessful robot grasps, or issues related to false positives or negatives when monitoring the activities carried out by human operators, the inherent flexibility of \textsc{ConcHRC} allows human operators to intervene and manage these issues. 
This is even more relevant considering that our current setup does not focus on such a robustness level. 

The second is the insight that using parallel instances of AND/OR graph representation layers seems to be more efficient with respect to an equivalent, common, single instance model. 
We observed that the adoption of \textsc{ConcHRC} reduces the overall idle time considerably.
This is an obvious consequence of the fact that the total time needed for a multi human-robot collaboration process to conclude is determined by the maximum one associated with the longest execution branch in the graph. 
On the contrary, if the HRC process were implemented as a single, non concurrent, model, then the total time would correspond to the sum of all times associated with single cooperation paths. 
As an example, in our scenario \textsc{ConcHRC} allows for a total collaboration time equal to $310.19$ $s$, whereas an equivalent implementation using \textsc{FlexHRC} the total collaboration time can be up to $866.94$ $s$.


\section{Conclusions}
\label{sec5}

In this paper we present and discuss \textsc{ConcHRC}, a framework aimed at modelling multi human-robot collaboration processes. 
The framework builds upon \textsc{FlexHRC}, which did not consider concurrent task allocation and execution.
\textsc{ConcHRC} has been preliminary analysed in a use case related to defects inspection, where one human operator and four robot \textit{agents} are present.
Two general remarks can be done.
The first is a general robustness of the human-robot cooperation flow with respect to issues related to object grasping and manipulation, as well as the recognition of human actions.
The second, which is related to best practices in modelling the cooperation scenario, is a tendency towards minimising idle times.

Obviously enough, the work can be improved along many directions:
(i) evaluating the use of a scheduler instead of a set of concurrent planners, especially considering approaches based on Answer Set Programming or metaheuristics; 
(ii) the gesture recognition module, used to detect and classify human activities, may be improved allowing for models able to predict them.
These two aspects are subject of current work.





\bibliographystyle{IEEEtran}
\bibliography{bibs/IEEEexample, bibs/IEEEabrv}

\end{document}